\documentclass[11pt,a4paper]{article}
\usepackage[hyperref]{emnlp2020}
\usepackage{times}
\usepackage{latexsym}

\usepackage{microtype}

\usepackage{times}
\usepackage{latexsym}
\usepackage{mathrsfs}
\usepackage{bm}
\usepackage{amsfonts,amssymb}
\usepackage{booktabs}
\usepackage{algorithm}
\usepackage{algorithmic} 
\usepackage{graphicx}
\usepackage{color}
\usepackage{multirow}

\usepackage{microtype}
\usepackage{verbatim}
\aclfinalcopy 

\title{Distantly Supervised Relation Extraction in Federated Settings}

\author{
	Dianbo Sui$^{1,2}$,
	Yubo Chen$^1$,
    Kang Liu$^{1,2}$,
	Jun Zhao$^{1,2}$\\
	$^1$ National Laboratory of Pattern Recognition, Institute of Automation, \\
	Chinese Academy of Sciences, Beijing, 100190, China \\
	$^2$ University of Chinese Academy of Sciences, Beijing, 100049, China \\
	\{dianbo.sui, yubo.chen, kliu,  jzhao\}@nlpr.ia.ac.cn \\
}

\date{}

\begin{document}
\maketitle
\begin{abstract}
This paper investigates distantly supervised relation extraction in federated settings. Previous studies focus on distant supervision under the assumption of centralized training, which requires collecting texts from different platforms and storing them on one machine. However, centralized training is challenged by two issues, namely, data barriers and privacy protection, which make it almost impossible or cost-prohibitive to centralize data from multiple platforms. Therefore, it is worthy to investigate distant supervision in the federated learning paradigm, which decouples the model training from the need for direct access to the raw data. Overcoming label noise of distant supervision, however, becomes more difficult in federated settings, since the sentences containing the same entity pair may scatter around different platforms. In this paper, we propose a federated denoising framework to suppress label noise in federated settings. The core of this framework is a multiple instance learning based denoising method that is able to select reliable instances via cross-platform collaboration. Various experimental results on New York Times dataset and miRNA gene regulation relation dataset demonstrate the effectiveness of the proposed method. 
\end{abstract}

\section{Introduction}
Relation extraction (RE) is a fundamental task for
knowledge base (KB) construction. It aims to mine factual knowledge from free texts by labeling relations between entity mentions.
Most existing supervised RE systems, such as \citet{zeng-etal-2014-relation,zhang2015relation,wang2016relation,zhou2016attention}, rely on large-scale manually annotated training data, which is labor-intensive and time-consuming. 

To ease the reliance on annotated data, \citet{mintz2009distant} proposed the distant supervision to automatically generate training data by aligning a KB and unlabeled texts. The key assumption of distant supervision is that if two entities have a relation in the KB, then all sentences that mention these two entities will express this relation. A set of sentences containing the same entity pair is called a bag. Although distant supervision can scale up training data, the automatic labelling inevitably accompanies with label noise, which means not all sentences that mention an entity pair can represent the relation between them. Training on such noisy data will hinder the performance of the RE model.

There is a rich literature on handling label noise in distant supervision, such as \citet{riedel2010modeling,hoffmann2011knowledge,zeng2015distant,lin2016neural,ye2019distant}. However, it should be noted that all these studies were conducted under the assumption of centralized training. In other words, all these approaches require centralizing texts from different platforms to one machine and then conduct training.  Centralized training faces two major challenges \cite{yang2019federated}. The first challenge is the data barrier, caused by the reluctance of data holders in most industries to share the underlying data. The second challenge is the fact that states across the world have been strengthening relevant laws in privacy protection, such as GDPR \footnote{\url{https://en.wikipedia.org/wiki/GDPR}}, which places a significant compliance burden on data collection. These two challenges make it almost  impossible or cost-prohibitive to integrate data from multiple platforms. Therefore, it is worthy to investigate distant supervision under 
the federated learning paradigm \cite{mcmahan2016communication}, which permits learning to be done while local data of each platform stays in its local environment.

Federated learning decouples the model training from the need for direct access to raw training data. The learning task is solved by a loose federation of platforms coordinated by a master server. Each platform has a local training dataset that is never uploaded to the master server. Instead, each platform  computes an update to the current global model maintained by the master server, and only this update is communicated between platforms and the master server. 

\begin{figure}[t] \includegraphics*[clip=true,width=0.49\textwidth,height=0.13\textheight]{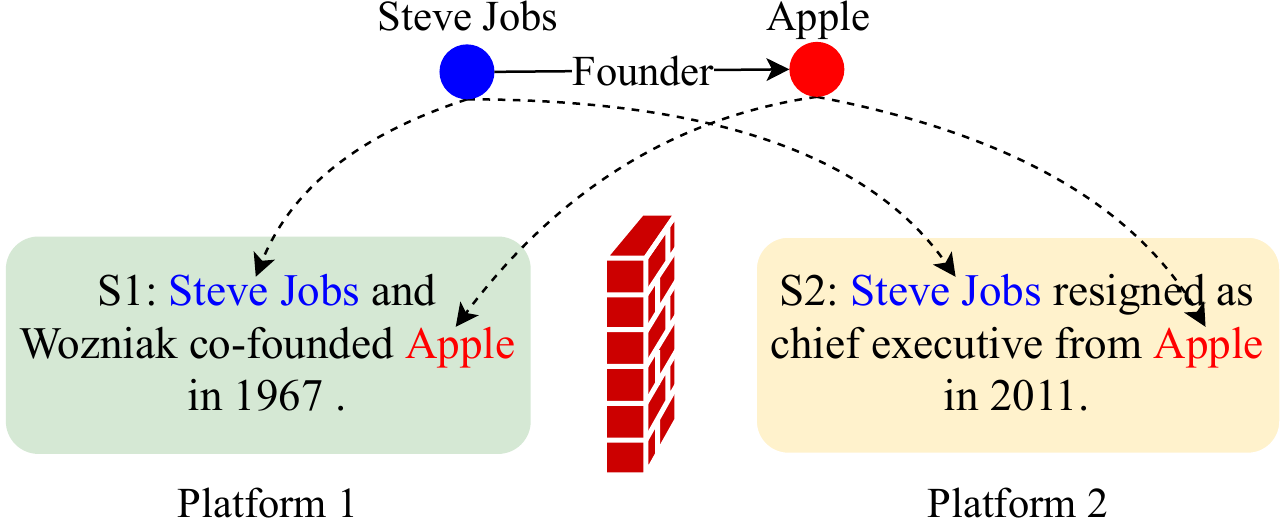}
  	\caption{An example of the sentences in a bag distributed on two platforms.} \label{introd}
  \end{figure}

Although in federated learning, texts from multiple platforms can be used to collaboratively train the model without considering data barriers and privacy issues, overcoming label noise of distant supervision becomes more arduous. In detail, the sentences in a bag may scatter around different platforms. As shown in Figure \ref{introd}, $S_1$ and $S_2$  contain the same entity pair (\textit{``Steve Jobs"}, \textit{``Apple"}) but are distributed on two platforms. $S_1$ is true positive while $S_2$ is a false positive instance, which does not express the \textit{``founder"} relation. Under the assumption of centralized training,  considering $S_1$ and $S_2$ simultaneously can easily denoise via only selecting $S_1$ \cite{zeng2015distant} or placing a small weight on $S_2$ \cite{lin2016neural,ye2019distant}.
However, due to data barriers or privacy issues, data exchange between platforms is prohibited. Without $S_1$, the false positive sentence $S_2$ is retained as training data instead of being removed as a noise. As a result, the local model in platform 2 is poisoned by the false positive instance $S_2$, which would in turn affect the global model.

To suppress label noise of distant supervision in federated settings, we propose a federated denoising framework in this paper. The core of this framework is a multiple instance learning (MIL) \cite{dietterich1997solving,maron1998framework} based denoising algorithm, called \textbf{Lazy MIL}, which is only executed at the beginning of each communication round and then would rest until the next round. Since the instances in a bag may scatter around different platforms, Lazy MIL algorithm coordinates multiple platforms to jointly select reliable instances without exposing underlying texts. Once instances have been selected, they would be used repeatedly to train local models until the end of this round. 

In summary, the contributions of this paper are

\begin{itemize}
    \item Considering data barriers and privacy protection, we investigate distant supervision under the federated learning paradigm, which decouples the model training from the need for direct access to the raw training data.
    \item To suppress label noise of distant supervision in federated settings, we present a multiple instance learning based denoising method, which can select reliable instances via cross-platform collaboration.
    \item The method yields promising results on two benchmarks datasets, and we perform various experiments to verify the effectiveness of the proposed method. The code will be released at \url{https://github.com/DianboWork/FedDS}.
\end{itemize}
\begin{figure*}[t]
  	\begin{center} \includegraphics*[clip=true,width=0.9375\textwidth,height=0.32\textheight]{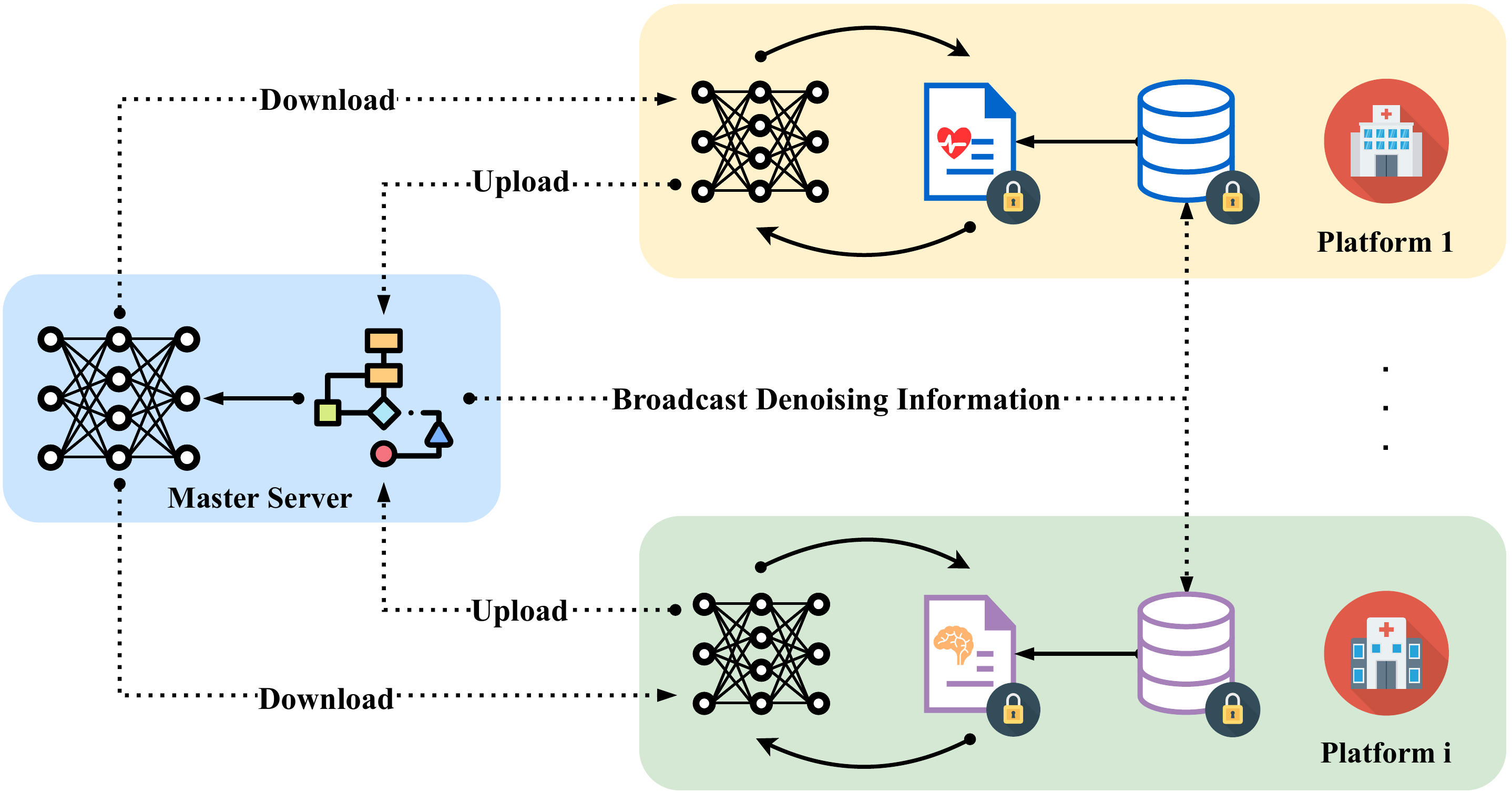}
  	\caption{The main architecture of federated denoising, where the dotted arrows represent the communication flow and the solid curve arrows indicate the training flow. At the beginning of each communication round, the global model parameters are distributed to activated platforms for initializing local models. Next, activated platforms receive the broadcast denoising information to select reliable training instances. Then, selected instances are used to train the local model at each platform. At the end of this round, all trained local model parameters are uploaded to the master server for updating the global model.} \label{fig}
  \end{center}
  \end{figure*}
\section{Related Work}
\label{related_work}
In this section, we will briefly review the recent progress in distantly supervised relation classification and existing studies on federated learning.
\subsection{Distantly Supervised Relation Extraction}
Relation extraction is a task of mining factual knowledge from free texts by labeling relations. To alleviate the dependence of supervised methods on annotated data, \citet{mintz2009distant} proposed distant supervision by using an existing knowledge base to automatically annotating large-scale datasets. However, distant supervision often suffers from label noise. To deal with label noise, most distantly supervised approaches \cite{riedel2010modeling,hoffmann2011knowledge,surdeanu2012multi,zeng2015distant,lin2016neural,ye2019distant} fall under the framework of multiple instance earning,  which assumes that at least one sentence expresses the relation in a bag. Our work is in line with this framework and, moreover, we extend this framework to federated settings. Another line of work aims to reduce label noise at sentence level prediction, some studies \cite{zeng2018large,feng2018reinforcement,qin2018dsgan,qin2018robust} use reinforcement learning or adversarial training to selects trustable relation labels by matching the predicted label of the learned model with distant supervision generated label. 
\subsection{Federated Learning}
Recently, federated learning \citep{mcmahan2016communication,konevcny2016federated1,konevcny2016federated2} has become a rapidly developing topic in the research community, since it provides a new communication-efficient way of learning models over a collection of highly distributed platforms while still preserving data privacy. According to distribution characteristics of the data, federated learning can be classified into horizontal federated learning, vertical federated learning and federated transfer learning \cite{yang2019federated}. This work is in line with the horizontal federated learning, where data sets share the same feature space but different in samples.

Federated learning has witnessed many successful applications in various fields. \citet{kim2017federated} introduced federated tensor factorization for computational phenotyping without sharing patient-level data. \citet{chen2018federated} combined federated learning with meta learning for the recommendation. \citet{2002.08562} presented federated pretraining of BERT model using clinical notes from multiple silos. \citet{ge2020fedner} applied federated learning to medical NER. In contrast to the previous work, we focus on applying federated learning to a noisy environment. To this end, we introduce a federated denoising framework.

\section{Federated Denoising Framework}
\label{methods}
\subsection{Task Definition}
In this paper, we focus on distant supervision in federated settings.
Define $K$ platforms $\{P_1, ...P_K\}$ with respective data $\{D_1, ...D_K\}$. Under the assumption of centralized training, each platform transfers its local data to a server, and the server will take the integrated data $D = D_1 \cup ... \cup  D_K$ to conduct training, while the task of distant supervision in federated settings requires any platform $P_i$ does not expose its data $D_i$ to others (including the server). In distant supervision, a knowledge base (KB) is required to automatically label the underlying texts. In this paper, we only focus on the data security of underlying texts, so the KB is public available for platforms. The issue of protecting the security of KB is beyond the scope of the current work.

To solve this task, we propose a federated denoising framework. The overall of this framework is shown in Figure \ref{fig} and the key components of this framework will be elaborated in the following section. Concretely, we first introduce the basic relation extractor in Section \ref{RE}, which is the network architecture shared by the global model and the local model. Then we present how to select reliable instance via cross-platform collaboration in Section \ref{lazyMIL}. Next, we describe how to use the selected instances to train the local model in Section \ref{local}. Finally, we present how to use federated averaging algorithm to update the global model in Section \ref{server}. 




\subsection{Relation Extractor}
\label{RE}
Following previous studies \cite{zeng2015distant,lin2016neural,ye2019distant}, we adopt the Piecewise Convolutional Neural Network (PCNN) as our relation extractor. 

Given a sentence $s$ and two entities within this sentence, we first split the sentence into tokens, and then each token $w_i$ is mapped into a dense word embedding $\mathbf{e}_i \in \mathbb{R}^{d_w}$. To specify entity pairs, the relative distances between the current token and the two entities are transformed into two positional features by looking up the position embedding matrices. Next, the token representation is represented as the concatenation of the word embedding and two positional features, and is fed into the convolutional neural network. Then, piecewise max pooling \cite{zeng2015distant} is employed to extract the high-level 
sentence representation from three segments of CNN outputs, and the boundaries of segments are determined by the positions of the two entities. After that, we apply a single fully connected layer to output the logit value $\mathbf{o}$.  Finally, the conditional probability of $j$-th relation is denoted as follows: 
\begin{equation}
    p(rel_j|s, \Theta) = \frac{exp(\mathbf{o}_j)}{\sum \limits_{i=1}^M exp(\mathbf{o}_i)}
    \label{eq1}
\end{equation}
where $\Theta$ is the model parameter and  M is the total number of relation. 


\subsection{Lazy Multiple Instance Learning}
\label{lazyMIL}
To avoid the local relation extractor being poisoned by false positive instances, we propose the lazy multiple instance learning (Lazy MIL), which can select reliable instances via cross-platform collaboration. The overall of Lazy MIL is illustrated in Algorithm \ref{alg1}.
\begin{algorithm}[ht]
  \caption{Lazy Multiple Instance Learning}
  \label{alg1}
  \begin{algorithmic}
    \STATE \textbf{Input}: $\Theta$ is the global model parameters, and $A$ is the set of activated platforms. 
     \STATE \textbf{Output}: \textit{V} is a dictionary about denoising information 

\STATE \textbf{Function} Lazy\_MIL ($\Theta$, $A$)
\STATE \quad Define a dictionary on the server, named \textit{V} 
\STATE \quad Distribute $\Theta$ to each platform in $A$
\STATE \quad \textsl{// Run on activated platforms}
\STATE \quad \textbf{for} each platform $i \in A$ \textbf{in parallel do}
\STATE \quad \quad \textbf{for} each triple $(h,r,t)$ in KB \textbf{do}
\STATE \quad \quad \quad \textbf{for} each sentence $s^i_z$ in bag $b^i$ \textbf{do}
\STATE \quad \quad \quad \quad Compute $p(r|s^i_z, \Theta)$ 
\STATE \quad \quad \quad $v^i, id^i \leftarrow \max_{z}(p(r|s^i_z, \Theta))$, $s^i_z \in b^i$
\STATE \quad \quad \quad Upload $[v^i, id^i, i]$ to the server
\STATE \quad \quad \quad \textsl{// Run on the master server}
\STATE \quad \quad \quad Add $[v^i, id^i, i]$ to  \textit{V}[$(h,r,t)$] 

\STATE \quad \textsl{// Run on the master server}
\STATE \quad \textbf{for} each key $(h,r,t)$ in \textit{V} \textbf{do}
\STATE \quad \quad \textsl{// The sorted list is in descending order}
\STATE \quad \quad tmp $\leftarrow$ sorted(\textit{V}[$(h,r,t)$], key = $\backslash$
\STATE \quad \quad \quad \textit{lambda} x:x[0], reverse=True)
\STATE \quad \quad \textit{V}[$(h,r,t)$] $\leftarrow$ tmp[0]
\STATE \quad \textbf{return} V
 \end{algorithmic}
\end{algorithm}

Suppose that there is a triple $(h, r, t)$ in the public KB \footnote{We assume there are triples with `NA' relation in the public KB. In other words, we treat `NA' as a normal relation.}, the set of sentences containing the head entity $h$ and tail entity $t$ is represented as $\{(s^1_1, s^1_2, ..., s^1_{n_1}), ..., (s^K_1, s^K_2, ..., s^K_{n_k})\}$, where $s^j_i$ indicates the $i$-th instance in the platform $j$. In the $q$-th communication round, assume that only platform $i$ and platform $j$ are activated. At the beginning of this round, the parameters of the global model $\Theta_q$ are distributed to the activated platforms $i$  and $j$ for initializing local models, which ensures that all activated local models share the same parameters in Lazy MIL. In platform $i$, the sentences in the set $(s^i_1, s^i_2, ..., s^i_{n_i})$ are fed into the local model to get conditional probabilities associated with the $r$ relation according to Equation \ref{eq1}, where $r$ is the predicate of the triple. The value $v^i$ and index $id^i$ of the instance with the maximum conditional probability associated with the $r$ relation is computed as follows\footnote{The max function returns the maximum value and the index location of the maximum value.}: 
\begin{equation}
    v^i, id^i = \max_{z}(p(r|s_z^i, \Theta_q)) \quad 1 \leq z \leq n_i 
\end{equation} 
After computation, platform $i$ uploads the value $v^i$ and index $id^i$ to the master server.  At the same time, the same procedure is performed on 
platform $j$, and the value $v^j$ and index $id^j$ are also uploaded to the server. 

The master server decides which local instance can be selected among all activated platforms based on the uploaded values. If $v^i > v^j$, then the $id^i$-th sentence in platform $i$ is selected as the reliable sentence that expresses this triple $(h,r,t)$ in this round. This decision, called denoising information, is broadcast to all activated platforms. Each activated platform selects reliable training instances from its local data according to this denoising information. Note that, since only
values and indices of conditional probabilities are uploaded to the master server, Lazy MIL does not leak the information of texts in each platform.
\subsection{Local Model Training}
\label{local}
After platform $i$ selects reliable instances from its local data $D_i$, the selected reliable instance set $D_i^\star$ is used for training the local relation extractor. We use the cross-entropy loss function to optimize parameters $\Theta_q$, which is defined as follows:
\begin{equation}
J(\Theta_q; D_i^\star) =  - \frac{1}{|D_i^\star|}\sum \limits_{u=1}^{|D_i^\star|} \log p(r_u|s^\star_u, \Theta_q)
\label{leq}
\end{equation}
where $s^\star_u$ indicates the $u$-th sentence in the selected reliable instance set $D_i^\star$. After training $E$ epochs on the selected reliable instance set, the trained parameters $\Theta_{q+1}^i$ are uploaded to the master server, where the superscript $i$ indicates the parameters are trained on platform $i$.

\subsection{Global Model Update}
\label{server}

Suppose $A_q$ is the set of activated platforms in the $q$-th communication round. After all activated platforms finish local training, the master server collects all trained parameters $\{\Theta_{q+1}^i | i \in A_q\}$ to update the global model. We define the goal of the global model as follows:
\begin{equation}
    \min \limits_{\Theta_{q}} \frac{1}{|A_q|}\sum \limits_{i \in A_q} J(\Theta_q; D^\star_i)
\end{equation}
where $J(\Theta_q; D^\star_i)$ is the local loss function for the platform $i$. Follow previous studies \cite{mcmahan2016communication}, we optimize this global objective function via taking an average of all trained parameters, which is shown as follows:
\begin{equation}
    \Theta_{q+1} = \frac{1}{|A_q|} \sum_{i \in A_q} \Theta_{q+1}^i
\end{equation}
where $\Theta_{q+1}^i$ is the optimal parameters obtained by minimizing the local loss function on the local data of platform $i$. Since all trained parameters from different platforms are aggregated together, the information of texts in each platform is hard to be inferred. Thus, texts in platforms are well-protected. Complete pseudo-code of this framework is given in  Algorithm \ref{alg2}. 
\begin{algorithm}[t]
  \caption{Federated Denoising Framework.}
  \label{alg2}
  \begin{algorithmic}
    \STATE \textbf{Hyperparameters:}
    \STATE \quad $K$ is the total number of platforms;
    \STATE \quad $C$ is the fraction of platforms;
    \STATE \quad $B$ is the local minibatch size;
    \STATE \quad $E$ is the number of local epochs;
    \STATE \quad $\eta$ is the learning rate.
    \end{algorithmic}
\begin{algorithmic}
\STATE\textbf{Master server executes:} 
      \STATE \quad Initialize $\Theta_0$
      \STATE \quad \textbf{for} each communication round $q$ = 0,1,... \textbf{do}
      \STATE \quad \quad \textsl{// Select activated platforms}
       \STATE \quad \quad $m$ $\leftarrow$ $\max$($C\times K$, 1) 
       \STATE \quad \quad $A_q\leftarrow$ (random set of $m$ platforms)
        \STATE \quad \quad \textsl{// Lazy MIL is defined in Algorithm \ref{alg1}}
       \STATE \quad \quad  \textit{V} $\leftarrow$ Lazy\_MIL($\Theta_q$, $A_q$)
        \STATE \quad \quad Broadcast \textit{V} to each platform in $A_q$
       \STATE \quad \quad \textbf{for} each platform $i \in A_q$ \textbf{in parallel do}
       \STATE\quad \quad \quad  $\Theta_{q+1}^k\leftarrow$ Local\_Training($i$, $\Theta_q$)
       \STATE\quad \quad \quad Upload $\Theta_{q+1}^k$ to the server
    \STATE\quad \quad  $\Theta_{q+1}\leftarrow \frac{1}{|A_q|} \sum_{i \in A_q} \Theta_{q+1}^i$
\STATE \quad
\STATE \textbf{Function} Local\_Training($i$, $\Theta_q$): 
\STATE \quad \textsl{// Run on platform $i$}
\STATE \quad Generate denoised dataset $D_i^\star$ from $D_i$ based
\STATE \quad \quad on the denoising information \textit{V}
\STATE \quad  $\mathcal{B} \leftarrow$ (split $D_i^\star$ into batches of size $B$)
\STATE \quad \textbf{for} each local epoch $e$ from $1$ to $E$ \textbf{do}
\STATE  \quad \quad \textbf{for} batch $b \in \mathcal{B}$ \textbf{do}
\STATE  \quad \quad \quad \textsl{// $J$ is defined in Equation \ref{leq}}
\STATE  \quad \quad \quad $\Theta \leftarrow \Theta - \eta \nabla J(\Theta;b)$
\STATE  \quad \textbf{return} $\Theta$
 \end{algorithmic}
\end{algorithm}

\section{Experiments}
\label{exp}

\subsection{Datasets and Evaluation Metrics}
We conduct experiments on two public available distantly supervised relation extraction data, i.g., NYT 10 dataset  \citep{riedel2010modeling}\footnote{\url{https://github.com/thunlp/OpenNRE}} and miRNA gene regulation relation (MIRGENE) dataset \cite{li-etal-2017-noise}\footnote{\url{https://github.com/leebird/bionlp17}}, to investigate the effectiveness of our method.


\textbf{NYT 10} is a standard benchmark distantly supervised dataset in news domain. It was automatically generated by aligning Freebase relations with the New York Times corpus,
with the years 2005–2006 reserved for training and validation
and 2007 for testing. The training data  
contains 466,876 sentences, 251,928 entity pairs
and 16,444 relational facts. The test data contains
172,448 sentences, 96,678 entity pairs and 1,950
relational facts. There are 52 actual relations and a special relation NA for representing no relation between two entities. 

\textbf{MIRGENE} is a large biomedical with 172727 sentences in the training set and 1239 sentences in the test set, and is generated by aligning Tarbase and miRTarBase with the Medline abstract. An example is shown in the following: ``\textit{ \textbf{MicroRNA-223} regulates \textbf{FOXO1} expression and cell proliferation}", where MicroRNA-223 is a miRNA and FOXO1 is a gene. 

\textbf{Data Partitioning}. To study distant supervision in federated settings, we need to specify how the data is distributed over the platforms. In this paper, we focus on the IID situation, where the training data is shuffled and then partitioned into $K$ (the total number of platforms) platforms.

\textbf{Evaluation Metrics}. We evaluate our approach and baseline methods on the held-out test set of these two datasets. Precision-recall (PR) curves, area under curve (AUC) values and Precision@N (P@N) values are adopted as evaluation metrics in our experiments. 
\subsection{Experimental Settings}
For a fair comparison, we implemented our method and all baselines in the same experimental settings. We divide the hyperparameters into three parts, i.e., fixed hyperparameters, unfixed hyperparameters and federated hyperparameters. Fixed hyperparameters follow the hyperparameter settings in \citet{lin2016neural}, including
the 50-dimensional pretrained word embeddings for NYT, the 5-dimensional position embeddings, and CNN module which includes 230 filters with a window size of 3. For MIRGENE, 200-dimensional word embeddings pretrained on PubMed and MIMIC-III texts are used. The optimal unfixed hyperparameters are determined by a grid search, and the search space of unfixed hyperparameters is shown in Table \ref{hyperparamter}.  Federated hyperparameters include the total number of platforms $K$, the fraction of platforms $C$, the local minibatch size $B$, the number of local epochs $E$. All of these control the amount of computation. In the end-to-end comparison, we fix the $K$ to 100, $B$ to 32, $E$ to 3, and set the hyperparameter space of $C$ as \{0.1, 0.2, 0.5, 1\} following \cite{mcmahan2016communication}.  We use stochastic gradient descent as the local training optimizer and all experiments can be done by using a single GeForce GTX 1080 Ti.
\begin{table}[!t]
\centering
\scalebox{0.8}{
\begin{tabular}{c|c}
\hline
Hyperparameter      & Search Space               \\
\hline \hline
Learning Rate ($\eta$)       & 0.05, 0.08, 0.1,0.2 \\
Learning Rate Decay & 0.01, 0.05          \\
Dropout             & 0.1, 0.2, 0.5       \\
Weight Decay        & 1e-5, 1e-6         \\
\hline
\end{tabular}}
\caption{The search space of unfixed hyperparameter.}
	\label{hyperparamter}
\end{table}

\subsection{Baselines}
We compare our method with the following denoising baselines in federated settings: (1) \citet{zeng2015distant} proposed to leverage  multi-instance learning to choose the most reliable sentence as the bag representation, and we abbreviate this method as \textbf{ONE}; (2) \textbf{ATT} is proposed by \citet{lin2016neural}, which uses the attention mechanisms to place soft weights on a set
of noisy sentences and select samples; (3) \textbf{AVE} \cite{lin2016neural} is a naive version of ATT and represents each sentence set as the average vector of sentences inside the set; (4) \textbf{ATT\_RA} \cite{ye2019distant} is a variant of ATT,  which calculates the bag representations in a relation-aware way. For a fair comparison, we keep the other modules unchanged and only replace the denoising module in this work with the baseline models.

\begin{figure*}[t]
  	\begin{center} \includegraphics*[clip=true,width=1\textwidth,height=0.2\textheight]{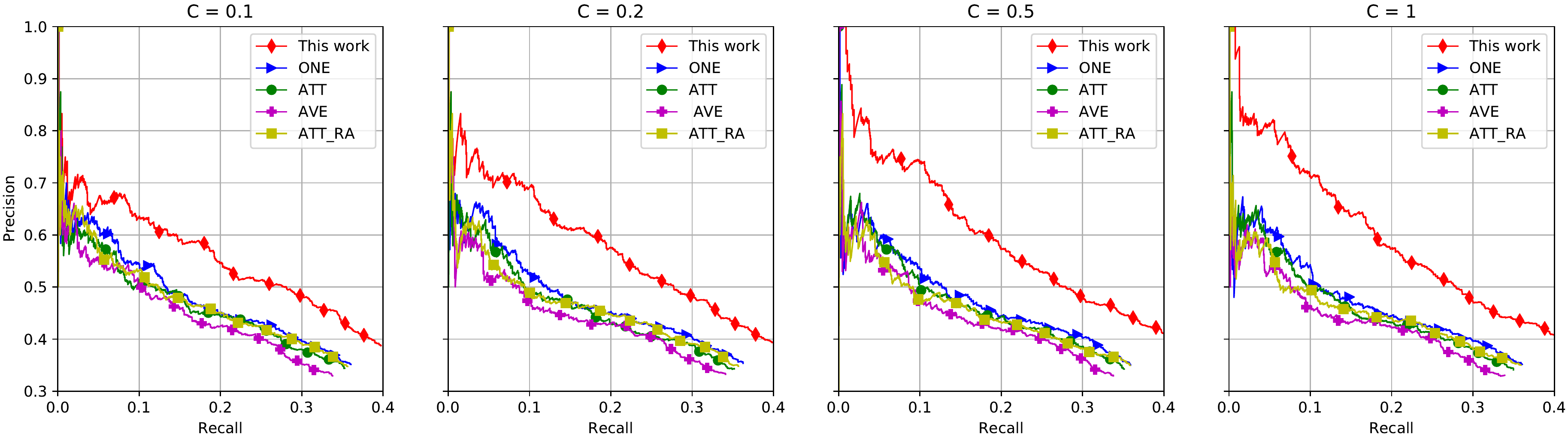}
  	\caption{Aggregate precision-recall curves on NYT 10 dataset, where $C$ is the fraction of platforms that are activated on each round.} \label{nyt}
  \end{center}
  \end{figure*}
\begin{table*}[h]
 \centering
\scalebox{0.6}{
\begin{tabular}{|c||cccc|cccc|cccc|cccc|}
\hline
  \multirow{2}*{\textbf{P@N(\%)}}       & \multicolumn{4}{c|}{C = 0.1}   & \multicolumn{4}{c|}{C = 0.2}   & \multicolumn{4}{c|}{C = 0.5}   & \multicolumn{4}{c|}{C = 1}     \\
\cline{2-17}
                   & P@100 & P@200 & P@300 & Mean  & P@100 & P@200 & P@300 & Mean  & P@100 & P@200 & P@300 & Mean  & P@100 & P@200 & P@300 & Mean  \\
\hline \hline
\textbf{AVG}       & 57    & 55         & 53    & 55    & 59    & 51.5       & 50.67 & 53.72 & 58    & 53.5       & 52.67 & 54.72 & 60    & 54       & 49.67 & 54.56 \\
\textbf{ONE}      & 63    & 60         & 54.67 & 59.22 & 66    & 58.5       & 55    & 59.83 & 65    & 59         & 55    & 59.67 & 62    & 60       & 56    & 59.33 \\
\textbf{ATT}       & 60    & 57         & 52.67 & 56.56 & 59    & 57         & 52.67 & 56.22 & 63    & 57.5       & 53.33 & 57.94 & 65    & 56.5     & 52.33 & 57.94 \\
\textbf{ATT\_RA} & 62    & 55.5       & 53.33 & 56.94 & 61    & 54         & 51    & 55.33 & 60    & 54.5       & 50.33 & 54.94 & 60    & 54.5     & 49    & 54.5  \\
\textbf{This work} & \textbf{69}    & \textbf{67}        & \textbf{63}    & \textbf{66.33} & \textbf{74}    & \textbf{70.5}       & \textbf{68.67} & \textbf{71.06} & \textbf{77}    & \textbf{74.5}       & \textbf{71.67} & \textbf{74.39} & \textbf{80}    & \textbf{75.5}     & \textbf{71.33} & \textbf{75.61}\\
\hline

\end{tabular}}
\caption{P@100, P@200, P@300 and the mean of them for each model in held-out evaluation on NYT 10 dataset.}
\label{PN_nyt}
\end{table*}  
\subsection{Results}
\subsubsection{Results on NYT 10}
We plot PR curves of all methods with the top 2000 points in Figure \ref{nyt}, present detailed precision values measured at different points along these curves in Table \ref{PN_nyt}, and show the AUC values of these curves in Table \ref{auc_nyt}. We find that: 
(1) Our method significantly outperforms all baselines in federated settings. We believe the reason is that our denoising method can hinder false positive instances from poisoning local models, which leads to a better performance of the global model. (2) $C$ is the fraction of platforms that are activated on each round, which controls the amount of multi-platform parallelism. With increasing platform parallelism, the performance of all baselines declines slightly while our method performs better. Intuitively, increasing platform parallelism is able to lead to better results, since involving more platforms in training can increase the likelihood that all sentences with the same entity pair appear simultaneously. However, due to lack of cross-platform collaboration, all baselines handle label noise only based on its own local data, which may hamper the performance. In contrast, our method selects reliable instances among all activated platforms, which can effectively reap the benefits of increasing platform parallelism. (3) Leveraging attention mechanisms to denoise, which is an effective solution in centralized settings, seems not to work in federated settings. Compared with centralized training, the sentences in a bag may scatter around different platforms in federated settings, so the number of the sentences with the same entity pair on a platform is small, which may lead to placing large attention weights on noisy sentences due to lack of inter-bag contrast. 

\subsubsection{Results on MIRGENE}
\begin{figure*}[t]
  	\begin{center} \includegraphics*[clip=true,width=1\textwidth,height=0.2\textheight]{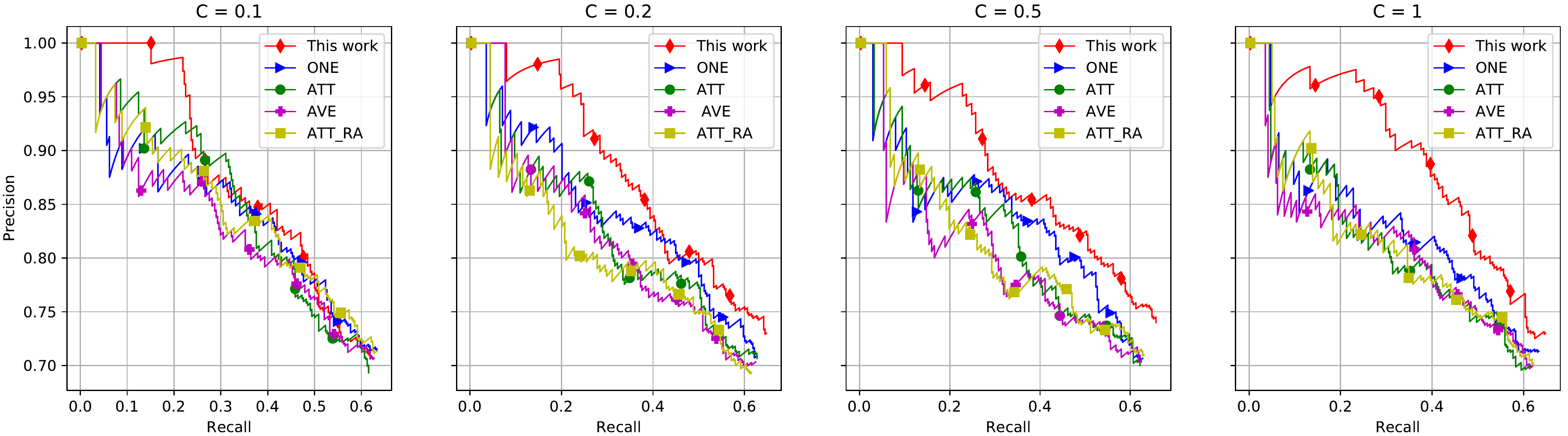}
  	\caption{Aggregate precision-recall curves on MIRGENE dataset, where $C$ is the fraction of platforms that are activated on each round.} \label{mirgene}
  \end{center}
  \end{figure*}

Figure \ref{mirgene}, Table \ref{PN_mirgene}, and Table \ref{auc_mirgene} show the comparison results in terms of  PR curves,  detailed precision values, and AUC values \footnote{We report the area under the curve of all points in the test set} respectively on  MIRGENE datasets. We notice that: (1) Our method achieves the best performance compared to all baselines, demonstrating the effectiveness of our denoising method. (2) With increasing platform parallelism (increasing $C$), baselines do not achieve better performance. In contrast, our method achieves better performance with the increase of parallelism. Unlike the significant performance improvement in NYT, the improvement of our method is not very obvious in MIRGENE. 
\begin{table}[t]
 \centering
\scalebox{0.90}{
\begin{tabular}{|c||cccc|}
\hline
\textbf{AUC}       & $C$ = 0.1 & $C$ = 0.2 & $C$ = 0.5 & $C$ = 1  \\
\hline \hline
AVG      & 0.1544  & 0.1531  & 0.1527  & 0.1503 \\
ONE       & 0.1747  & 0.1747  & 0.1725  & 0.1715 \\
ATT       & 0.1658  & 0.1642  & 0.1657  & 0.1631 \\
ATT\_RA   & 0.1695  & 0.1666  & 0.1647  & 0.1637 \\
This work & \textbf{0.2207}  & \textbf{0.2315}  & \textbf{0.2448}  & \textbf{0.2465}\\
\hline
\end{tabular}}
\caption{AUC values on NYT 10 dataset.}
\label{auc_nyt}
\end{table} 
We conjecture this is largely due to the characteristic of the dataset. Concretely, the average number of sentences containing same entity pairs with non-``NA" relation is about 8 in NYT 10 while the number is about 4 in MIRGENE, which means that with lower parallelism, there is a high probability that all sentences with the same entity pairs appear simultaneously in MIRGENE. 
  
 \begin{table*}[h]
 \centering
\scalebox{0.6}{
\begin{tabular}{|c||cccc|cccc|cccc|cccc|}
\hline
  \multirow{2}*{\textbf{P@N(\%)}}       & \multicolumn{4}{c|}{C = 0.1}   & \multicolumn{4}{c|}{C = 0.2}   & \multicolumn{4}{c|}{C = 0.5}   & \multicolumn{4}{c|}{C = 1}     \\
\cline{2-17}
                   & P@100 & P@200 & P@300 & Mean  & P@100 & P@200 & P@300 & Mean  & P@100 & P@200 & P@300 & Mean  & P@100 & P@200 & P@300 & Mean  \\
\hline \hline
\textbf{AVG}      & 87    & 78         & 70.67 & 78.56 & 85    & 76         & 70.33 & 77.11 & 84    & 75         & 70.33 & 76.44 & 82    & 77       & 70    & 76.33 \\
\textbf{ONE}      & 87    & 79.5       & \textbf{71.33} & 79.28 & 85    & 79.5       & 70.67 & 78.39 & 87    & 80         & 70.67 & 79.22 & 82    & 78.5     & 71.33 & 77.28 \\
\textbf{ATT}       & \textbf{89}    & 77.5       & 69.33 & 78.61 & 87    & 78         & 70.67 & 78.56 & 87    & 75         & 70    & 77.33 & 82    & 76       & 70    & 76    \\
\textbf{ATT\_RA}  & 86    & 77       & \textbf{71.33} & 78.11 & 80    & 76.5       & 69.33 & 75.28 & 83    & 77         & 71    & 77    & 83    & 76       & 70    & 76.33 \\
\textbf{This work} & \textbf{89}    & \textbf{80}         & 70.67 & \textbf{79.89} & \textbf{91}    & \textbf{80.5}       & \textbf{73}    & \textbf{81.5}  & \textbf{92}    & \textbf{82.5}& \textbf{74}    & \textbf{82.83} & \textbf{95}    & \textbf{82}     & \textbf{73}    & \textbf{83.3} \\\hline
\end{tabular}}
\caption{P@100, P@200, P@300 and the mean of them for each model in held-out evaluation on MIRGENE dataset.}
\label{PN_mirgene}
\end{table*} 
\begin{table}[t]
 \centering
\scalebox{0.90}{
\begin{tabular}{|c||cccc|}
\hline
\textbf{AUC}       & $C$ = 0.1 & $C$ = 0.2 & $C$ = 0.5 & $C$ = 1  \\
\hline \hline
AVG      & 0.7577  & 0.7491  & 0.7431  & 0.7432 \\
ONE       & 0.7705  & 0.7649  & 0.7626  & 0.757 \\
ATT       & 0.7696  & 0.7528  & 0.7516  & 0.7483 \\
ATT\_RA   & 0.7597  & 0.7448  & 0.7484  & 0.7493  \\
This work & \textbf{0.7893}  & \textbf{0.7923}  & \textbf{0.7946}  & \textbf{0.7966}\\
\hline
\end{tabular}}
\caption{AUC values on MIRGENE dataset.}
\label{auc_mirgene}
\end{table} 

\subsection{Savings of Local Computation}
\begin{figure}[t]
  	\begin{center} \includegraphics*[clip=true,width=0.44\textwidth,height=0.18\textheight]{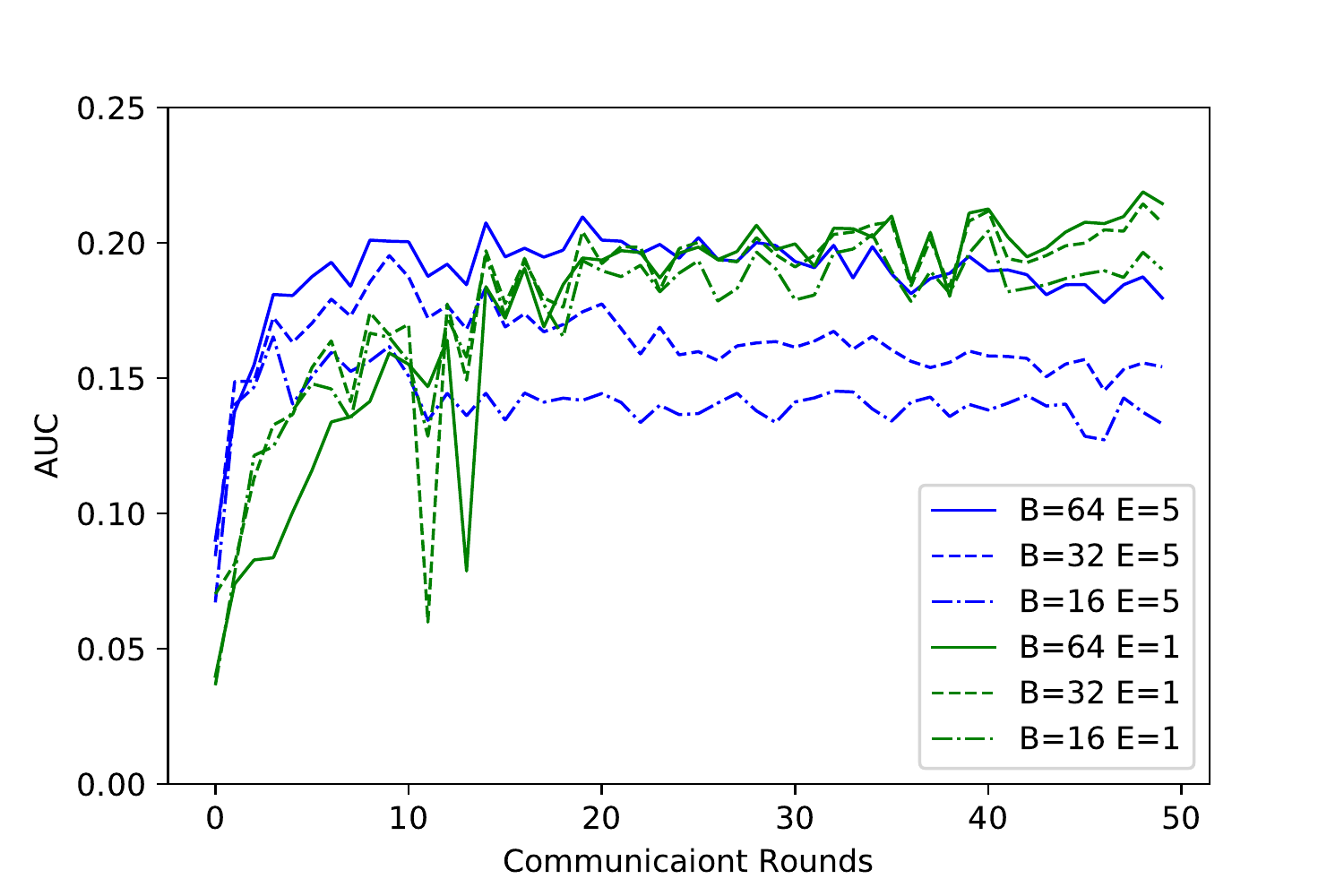}
  	\caption{AUC values vs. communication rounds on NYT data with different $E$ (the number of local epochs) and $B$ (the local minibatch size).} \label{batch_epoch}
  \end{center}
  \end{figure}

In the real-world scenario, platforms are controlled by data holders or users, which require conducting local training with the least computation cost. Therefore, we investigate the impact of varying the number of local updates in this section. The number of local updates is given by $E\frac{|D^*_i|}{B}$, where $|D^*_i|$ is the size of the denoised dataset in platform $i$ at a round, $B$ is the local minibatch size and $E$ is the number of local epochs. Increasing $B$, decreasing $E$, or both will reduce computation on each round. We fix $C$ to 0.1 and only $B$ and $E$ are varied in this section.\footnote{The lr, lr decay, weight decay and dropout are fix to is 0.1, 0.01, 1e-5 and 0.1 respectively, which are not the optimal hyperparameters for most experiments}. The results are shown in Figure \ref{batch_epoch}. We find that: (1) When setting $B$ to 64 and $E$ to 1, our method achieves the best AUC value. In this case, the number of local updates is the least. (2) Increasing the local minibatch $B$ may improve the performance. (3) Increasing the local epoch $E$ can make training more stable and speed up converge, but may not make the global model converge to a higher level of AUC value. These findings are in line with \citet{mcmahan2016communication}, which shows it may hurt performance when over-optimize on the local dataset. We also present the results of other baselines in the Appendix due to page limits.

\section{Conclusion}
\label{conclusion}
Due to data barriers and privacy protection, it is almost impossible or cost-prohibitive to integrate the data from multiple platforms. In this paper, we investigate distant supervision under the federated learning paradigm, which permits learning to be done while data stays in its local environment. 
To suppress label noise in federated settings, we propose a federated denoising framework, which can select reliable instances via cross platform collaboration. Extensive experiments on two datasets have demonstrated the effectiveness of our model.
\bibliography{emnlp2020}

\begin{thebibliography}{26}
\expandafter\ifx\csname natexlab\endcsname\relax\def\natexlab#1{#1}\fi

\bibitem[{Chen et~al.(2018)Chen, Dong, Li, and He}]{chen2018federated}
Fei Chen, Zhenhua Dong, Zhenguo Li, and Xiuqiang He. 2018.
\newblock Federated meta-learning for recommendation.
\newblock \emph{arXiv preprint arXiv:1802.07876}.

\bibitem[{Dietterich et~al.(1997)Dietterich, Lathrop, and
  Lozano-P{\'e}rez}]{dietterich1997solving}
Thomas~G Dietterich, Richard~H Lathrop, and Tom{\'a}s Lozano-P{\'e}rez. 1997.
\newblock Solving the multiple instance problem with axis-parallel rectangles.
\newblock \emph{Artificial intelligence}, 89(1-2):31--71.

\bibitem[{Feng et~al.(2018)Feng, Huang, Zhao, Yang, and
  Zhu}]{feng2018reinforcement}
Jun Feng, Minlie Huang, Li~Zhao, Yang Yang, and Xiaoyan Zhu. 2018.
\newblock Reinforcement learning for relation classification from noisy data.
\newblock In \emph{Thirty-Second AAAI Conference on Artificial Intelligence}.

\bibitem[{Ge et~al.(2020)Ge, Wu, Wu, Qi, Huang, and Xie}]{ge2020fedner}
Suyu Ge, Fangzhao Wu, Chuhan Wu, Tao Qi, Yongfeng Huang, and Xing Xie. 2020.
\newblock Fedner: Medical named entity recognition with federated learning.
\newblock \emph{arXiv preprint arXiv:2003.09288}.

\bibitem[{Hoffmann et~al.(2011)Hoffmann, Zhang, Ling, Zettlemoyer, and
  Weld}]{hoffmann2011knowledge}
Raphael Hoffmann, Congle Zhang, Xiao Ling, Luke Zettlemoyer, and Daniel~S Weld.
  2011.
\newblock Knowledge-based weak supervision for information extraction of
  overlapping relations.
\newblock In \emph{Proceedings of the 49th Annual Meeting of the Association
  for Computational Linguistics: Human Language Technologies-Volume 1}, pages
  541--550. Association for Computational Linguistics.

\bibitem[{Kim et~al.(2017)Kim, Sun, Yu, and Jiang}]{kim2017federated}
Yejin Kim, Jimeng Sun, Hwanjo Yu, and Xiaoqian Jiang. 2017.
\newblock Federated tensor factorization for computational phenotyping.
\newblock In \emph{Proceedings of the 23rd ACM SIGKDD International Conference
  on Knowledge Discovery and Data Mining}, pages 887--895. ACM.

\bibitem[{Kone{\v{c}}n{\`y} et~al.(2016{\natexlab{a}})Kone{\v{c}}n{\`y},
  McMahan, Ramage, and Richt{\'a}rik}]{konevcny2016federated1}
Jakub Kone{\v{c}}n{\`y}, H~Brendan McMahan, Daniel Ramage, and Peter
  Richt{\'a}rik. 2016{\natexlab{a}}.
\newblock Federated optimization: Distributed machine learning for on-device
  intelligence.
\newblock \emph{arXiv preprint arXiv:1610.02527}.

\bibitem[{Kone{\v{c}}n{\`y} et~al.(2016{\natexlab{b}})Kone{\v{c}}n{\`y},
  McMahan, Yu, Richt{\'a}rik, Suresh, and Bacon}]{konevcny2016federated2}
Jakub Kone{\v{c}}n{\`y}, H~Brendan McMahan, Felix~X Yu, Peter Richt{\'a}rik,
  Ananda~Theertha Suresh, and Dave Bacon. 2016{\natexlab{b}}.
\newblock Federated learning: Strategies for improving communication
  efficiency.
\newblock \emph{arXiv preprint arXiv:1610.05492}.

\bibitem[{Li et~al.(2017)Li, Wu, and Vijay-Shanker}]{li-etal-2017-noise}
Gang Li, Cathy Wu, and K.~Vijay-Shanker. 2017.
\newblock \href {https://doi.org/10.18653/v1/W17-2323} {Noise reduction methods
  for distantly supervised biomedical relation extraction}.
\newblock In \emph{{B}io{NLP} 2017}, pages 184--193, Vancouver, Canada,.
  Association for Computational Linguistics.

\bibitem[{Lin et~al.(2016)Lin, Shen, Liu, Luan, and Sun}]{lin2016neural}
Yankai Lin, Shiqi Shen, Zhiyuan Liu, Huanbo Luan, and Maosong Sun. 2016.
\newblock Neural relation extraction with selective attention over instances.
\newblock In \emph{Proceedings of the 54th Annual Meeting of the Association
  for Computational Linguistics (Volume 1: Long Papers)}, pages 2124--2133.

\bibitem[{Liu and Miller(2020)}]{2002.08562}
Dianbo Liu and Tim Miller. 2020.
\newblock Federated pretraining and fine tuning of bert using clinical notes
  from multiple silos.
\newblock \emph{arXiv preprint arXiv:2002.08562}.

\bibitem[{Maron and Lozano-P{\'e}rez(1998)}]{maron1998framework}
Oded Maron and Tom{\'a}s Lozano-P{\'e}rez. 1998.
\newblock A framework for multiple-instance learning.
\newblock In \emph{Advances in neural information processing systems}, pages
  570--576.

\bibitem[{McMahan et~al.(2016)McMahan, Moore, Ramage, Hampson
  et~al.}]{mcmahan2016communication}
H~Brendan McMahan, Eider Moore, Daniel Ramage, Seth Hampson, et~al. 2016.
\newblock Communication-efficient learning of deep networks from decentralized
  data.
\newblock \emph{arXiv preprint arXiv:1602.05629}.

\bibitem[{Mintz et~al.(2009)Mintz, Bills, Snow, and
  Jurafsky}]{mintz2009distant}
Mike Mintz, Steven Bills, Rion Snow, and Dan Jurafsky. 2009.
\newblock Distant supervision for relation extraction without labeled data.
\newblock In \emph{Proceedings of the Joint Conference of the 47th Annual
  Meeting of the ACL and the 4th International Joint Conference on Natural
  Language Processing of the AFNLP}, pages 1003--1011.

\bibitem[{Qin et~al.(2018{\natexlab{a}})Qin, Xu, and Wang}]{qin2018dsgan}
Pengda Qin, Weiran Xu, and William~Yang Wang. 2018{\natexlab{a}}.
\newblock Dsgan: Generative adversarial training for distant supervision
  relation extraction.
\newblock In \emph{Proceedings of the 56th Annual Meeting of the Association
  for Computational Linguistics (Volume 1: Long Papers)}, pages 496--505.

\bibitem[{Qin et~al.(2018{\natexlab{b}})Qin, Xu, and Wang}]{qin2018robust}
Pengda Qin, Weiran Xu, and William~Yang Wang. 2018{\natexlab{b}}.
\newblock Robust distant supervision relation extraction via deep reinforcement
  learning.
\newblock In \emph{Proceedings of the 56th Annual Meeting of the Association
  for Computational Linguistics (Volume 1: Long Papers)}, pages 2137--2147.

\bibitem[{Riedel et~al.(2010)Riedel, Yao, and McCallum}]{riedel2010modeling}
Sebastian Riedel, Limin Yao, and Andrew McCallum. 2010.
\newblock Modeling relations and their mentions without labeled text.
\newblock In \emph{Joint European Conference on Machine Learning and Knowledge
  Discovery in Databases}, pages 148--163. Springer.

\bibitem[{Surdeanu et~al.(2012)Surdeanu, Tibshirani, Nallapati, and
  Manning}]{surdeanu2012multi}
Mihai Surdeanu, Julie Tibshirani, Ramesh Nallapati, and Christopher~D. Manning.
  2012.
\newblock \href {https://www.aclweb.org/anthology/D12-1042} {Multi-instance
  multi-label learning for relation extraction}.
\newblock In \emph{Proceedings of the 2012 Joint Conference on Empirical
  Methods in Natural Language Processing and Computational Natural Language
  Learning}, pages 455--465, Jeju Island, Korea. Association for Computational
  Linguistics.

\bibitem[{Wang et~al.(2016)Wang, Cao, De~Melo, and Liu}]{wang2016relation}
Linlin Wang, Zhu Cao, Gerard De~Melo, and Zhiyuan Liu. 2016.
\newblock Relation classification via multi-level attention cnns.
\newblock In \emph{Proceedings of the 54th Annual Meeting of the Association
  for Computational Linguistics (Volume 1: Long Papers)}, pages 1298--1307.

\bibitem[{Yang et~al.(2019)Yang, Liu, Chen, and Tong}]{yang2019federated}
Qiang Yang, Yang Liu, Tianjian Chen, and Yongxin Tong. 2019.
\newblock Federated machine learning: Concept and applications.
\newblock \emph{ACM Transactions on Intelligent Systems and Technology (TIST)},
  10(2):1--19.

\bibitem[{Ye and Ling(2019)}]{ye2019distant}
Zhi-Xiu Ye and Zhen-Hua Ling. 2019.
\newblock Distant supervision relation extraction with intra-bag and inter-bag
  attentions.
\newblock In \emph{Proceedings of the 2019 Conference of the North American
  Chapter of the Association for Computational Linguistics: Human Language
  Technologies, Volume 1 (Long and Short Papers)}, pages 2810--2819.

\bibitem[{Zeng et~al.(2015)Zeng, Liu, Chen, and Zhao}]{zeng2015distant}
Daojian Zeng, Kang Liu, Yubo Chen, and Jun Zhao. 2015.
\newblock Distant supervision for relation extraction via piecewise
  convolutional neural networks.
\newblock In \emph{Proceedings of the 2015 conference on empirical methods in
  natural language processing}, pages 1753--1762.

\bibitem[{Zeng et~al.(2014)Zeng, Liu, Lai, Zhou, and
  Zhao}]{zeng-etal-2014-relation}
Daojian Zeng, Kang Liu, Siwei Lai, Guangyou Zhou, and Jun Zhao. 2014.
\newblock Relation classification via convolutional deep neural network.
\newblock In \emph{Proceedings of {COLING} 2014, the 25th International
  Conference on Computational Linguistics: Technical Papers}, pages 2335--2344,
  Dublin, Ireland. Dublin City University and Association for Computational
  Linguistics.

\bibitem[{Zeng et~al.(2018)Zeng, He, Liu, and Zhao}]{zeng2018large}
Xiangrong Zeng, Shizhu He, Kang Liu, and Jun Zhao. 2018.
\newblock Large scaled relation extraction with reinforcement learning.
\newblock In \emph{Thirty-Second AAAI Conference on Artificial Intelligence}.

\bibitem[{Zhang and Wang(2015)}]{zhang2015relation}
Dongxu Zhang and Dong Wang. 2015.
\newblock Relation classification via recurrent neural network.
\newblock \emph{arXiv preprint arXiv:1508.01006}.

\bibitem[{Zhou et~al.(2016)Zhou, Shi, Tian, Qi, Li, Hao, and
  Xu}]{zhou2016attention}
Peng Zhou, Wei Shi, Jun Tian, Zhenyu Qi, Bingchen Li, Hongwei Hao, and Bo~Xu.
  2016.
\newblock Attention-based bidirectional long short-term memory networks for
  relation classification.
\newblock In \emph{Proceedings of the 54th annual meeting of the association
  for computational linguistics (volume 2: Short papers)}, pages 207--212.

\end{thebibliography}
\bibliographystyle{acl_natbib}

\end{document}